\documentclass[sigconf]{acmart}

\usepackage[noabbrev]{cleveref}
\usepackage{caption}
\usepackage{subcaption}
\usepackage{algorithm}
\usepackage{algpseudocode}
\usepackage{enumitem}
\usepackage{xcolor}
\usepackage{balance}
\usepackage{epstopdf}
\AtBeginDocument{%
    \providecommand\BibTeX{{%
        Bib\TeX}}}

\copyrightyear{2023}
\acmYear{2023}
\setcopyright{rightsretained}
\acmConference[HRI '23]{Proceedings of the 2023 ACM/IEEE International Conference on Human-Robot Interaction}{March 13--16, 2023}{Stockholm, Sweden}
\acmBooktitle{Proceedings of the 2023 ACM/IEEE International
Conference on Human-Robot Interaction (HRI '23), March 13--16, 2023, Stockholm, Sweden}
\acmDOI{10.1145/3568162.3577002}
\acmISBN{978-1-4503-9964-7/23/03}

\makeatletter
\gdef\@copyrightpermission{
  \begin{minipage}{0.3\columnwidth}
   \href{https://creativecommons.org/licenses/by/4.0/}{\includegraphics[width=0.90\textwidth]{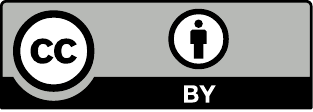}}
  \end{minipage}\hfill
  \begin{minipage}{0.7\columnwidth}
   \href{https://creativecommons.org/licenses/by/4.0/}{This work is licensed under a Creative Commons Attribution International 4.0 License.}
  \end{minipage}
  \vspace{5pt}
}
\makeatother

\acmPrice{15.00}

\newcommand{\checkstat}[1]{\textcolor{black}{#1}}
\newcommand{\hl}[1]{\textcolor{black}{#1}}

\begin{document}

\title{
 The Effect of Robot Skill Level and Communication in Rapid, Proximate Human-Robot Collaboration
}


\author{Kin Man Lee}
\authornote{Both authors contributed equally to this research.}
\email{klee863@gatech.edu}
\affiliation{
  \institution{Georgia Institute of Technology}
  \city{Atlanta}
  \state{Georgia}
  \country{USA}
}

\author{Arjun Krishna}
\authornotemark[1]
\email{akrishna49@gatech.edu}
\affiliation{
  \institution{Georgia Institute of Technology}
  \city{Atlanta}
  \state{Georgia}
  \country{USA}
}

\author{Zulfiqar Zaidi}
\email{zzaidi8@gatech.edu}
\affiliation{
  \institution{Georgia Institute of Technology}
  \city{Atlanta}
  \state{Georgia}
  \country{USA}
}

\author{Rohan Paleja}
\email{rpaleja3@gatech.edu}
\affiliation{
  \institution{Georgia Institute of Technology}
  \city{Atlanta}
  \state{Georgia}
  \country{USA}
}

\author{Letian Chen}
\email{letian.chen@gatech.edu}
\affiliation{
  \institution{Georgia Institute of Technology}
  \city{Atlanta}
  \state{Georgia}
  \country{USA}
}

\author{Erin Hedlund-Botti}
\email{erin.botti@gatech.edu}
\affiliation{
  \institution{Georgia Institute of Technology}
  \city{Atlanta}
  \state{Georgia}
  \country{USA}
}

\author{Mariah Schrum}
\email{mschrum3@gatech.edu}
\affiliation{
  \institution{Georgia Institute of Technology}
  \city{Atlanta}
  \state{Georgia}
  \country{USA}
}

\author{Matthew Gombolay}
\email{matthew.gombolay@cc.gatech.edu}
\affiliation{
  \institution{Georgia Institute of Technology}
  \city{Atlanta}
  \state{Georgia}
  \country{USA}
}

\renewcommand{\shortauthors}{Kin Man Lee et al.}

\newcommand\theme[1]{\noindent\textcolor{violet}{\underline{Theme:} \textit{#1}}\\}
\newcommand\kml[1]{\textcolor{orange}{(#1)}}
\newcommand\ark[1]{\textcolor{blue}{([ark]: #1)}}

\newcommand\suggest[1]{\textcolor{red}{#1}}

\newcommand\zone[1]{Zone {\it #1}}
\newcommand\RQ[1]{({\bf #1})}

\begin{abstract}
As high-speed, agile robots become more commonplace, these robots will have the potential to better aid and collaborate with humans. However, due to the increased agility and functionality of these robots, close collaboration with humans can create safety concerns that alter team dynamics and degrade task performance. In this work, we aim to enable the deployment of safe and trustworthy agile robots that operate in proximity with humans. We do so by 1) Proposing a novel human-robot doubles table tennis scenario to serve as a testbed for studying agile, proximate human-robot collaboration and 2) Conducting a user-study to understand how attributes of the robot (e.g., robot competency or capacity to communicate) impact team dynamics, perceived safety, and perceived trust, and how these latent factors affect human-robot collaboration (HRC) performance. We find that robot competency significantly increases perceived trust ($p<.001$), extending skill-to-trust assessments in prior studies to agile, proximate HRC. Furthermore, interestingly, we find that when the robot vocalizes its intention to perform a task, it results in a significant decrease in team performance ($p=.037$) and perceived safety of the system ($p=.009$).

\end{abstract}




\maketitle
\section{Introduction}
The deployment of collaborative robots (i.e., ``cobots'') holds the promise of increasing productivity and enhancing safety. These robots are expected to collaborate in proximity with human workers as opposed to the caged robots typically encountered within today's manufacturing industry \cite{Villani2018SurveyOH}. This paradigm shift allows for cobots to work with human workers in new ways, such as mixed-initiative teaming \cite{mingyue2018human,Gombolay2017ComputationalDO,jiang2015mixed}, to increase productivity. In mixed-initiative settings, human workers and robots are co-located in a shared workspace and must dynamically decide who should accomplish shared tasks. This online reasoning in human-robot collaboration provides a more natural and fluent interaction scheme than  a strict, pre-planned allocation of work~\cite{Angleraud2021CoordinatingST}. 

However, achieving fluency in mixed-initiative teaming is challenging as identifying precisely who should accomplish a specific task is computationally challenging for human and robot alike \cite{Ravichandar2020STRATAUF}, and inferring human intent online from dynamic motion remains an open research problem requiring domain expertise for algorithm design~\cite{RAVICHANDAR2017217}. Enabling robots to convey intent is likewise an unsolved design problem~\cite{dragan2015effects, Paleja2022TheUO}. Furthermore, as robots gain functionality and increased agility, close collaboration can create safety concerns that may alter the team dynamic, affecting human behavior and performance due to latent variables, such as perceived safety and trust~\cite{lasota2017survey,lasota2014toward}. 
In this work, we are interested in understanding how attributes of the robot (e.g., skill-level or capacity to communicate) impact team dynamics, perceived safety, and perceived trust, and how these latent factors affect human-robot collaboration (HRC) performance. Insight into these relationships will inform how to best design robots that work in proximity to humans. 

We are specifically interested in the development of agile robotics for human-robot collaboration. Agile robots are those that are, by definition, capable of demonstrating agile behaviors in dynamic settings~\cite{dong_catch_2020,yang-varsm2022}. These robots are designed to closely emulate human-level abilities across tasks such as sports ~\cite{yang-varsm2022,buchler2022learning}, cooking \cite{pancakeflippingrobot,cookingrobot,kitchen_robots}, and mobile manipulation \cite{RoboImitationPeng20,survey_humanoids}. 
Agile robotics holds the promise of low-latency decision-making and high-speed robot maneuvers. Collaboration in proximity to humans presents a novel challenge for agile robotics, as we would like to take advantage of high-speed manipulation for effective human-robot collaboration. 

Fast moving collaborative robots in proximity to humans could potentially induce fear, anxiety, and stress because the robot's intention may be unclear, the behaviors exhibited by the robot could be unpredictable, and the risk of the human coming to harm is objectively higher~\cite{stress_around_robots,Salvini2022,Murashov2016-vi,saferobotmotionspeed_industrialrobots}. Additionally, humans have less time to react, exacerbating the concerns of safety and trust~\cite{stress_around_robots,Story2022,Butler2001,SHIBATA1998483}. 
In this work, we choose the domain of table-tennis doubles for our human-subjects experiment in order to gain insights into the design of agile, collaborative robots. Here, we do not follow the typical table-tennis doubles rule, where agents must alternate hitting the ball; instead, we frame our domain as a mixed-initiative human-robot teaming task (i.e., either player can return the ball). Our modified game introduces the opportunity for players to reason about who should hit the ball, whether and how to communicate intent, and a possible physical collision if the human and robot are not in sync. Our task meets the following criterion: (1) Low-reaction time decisions from humans and robots, and (2) Shared collaboration space where either agent can successfully accomplish the task. In this environment, we evaluate how an agile robot, varying in skill level and endowed with a deliberative voice-communication channel is able to team with human end-users. We place specific emphasis on understanding how the robot's competence, voice profile, and the capacity to communicate affect perceived safety, and trust in agile, proximate human-robot collaboration. We note that we include as an independent variable the sex of the voice (i.e., male vs.~female) due to (conflicting) prior work in human factors on sex-binary voices for cockpit automation indicating that the sound of female vs.~male voice can have differing impacts on audibility, perceived assertiveness, and reaction time and quality~\cite{edworthy2003use,arrabito2009effects}.
 

While we study these factors in an \emph{ad hoc} table-tennis doubles task, the outcomes of this study are potentially applicable to settings where humans and agile-robots have to work together in a shared space.
 Applications include collaborative manufacturing robots, assistive home robots, athletic robots for sports and recreation, etc. Our work is the first to explore these factors in an athletic collaborative human-robot task and we hope this work inspires more research to enable the deployment of safe and trustworthy agile robots that operate in proximity with humans.
We summarize our key contributions as follows: 
\begin{enumerate}
    \item \hl{We develop a novel table tennis system for safe and dynamic HRC by extending probabilistic movement primitives~\cite{gomez2016using} for motor control and incorporating a safety subsystem that uses unintrusive, camera-based human pose tracking.}
    \item \hl{We conduct a user study and find that fast, verbal communication to convey intent reduces overall team performance ($p=.037$) and perceived safety  of the robot ($p=.009$). }
    \item \hl{We find that participants rarely utilized verbal signals to convey intent to the robot. Based on a small sample size of user feedback, this suggests that most felt more comfortable communicating to the robot implicitly with body motion in high-speed teaming tasks.}
\end{enumerate}



\begin{figure}
     \centering
     \includegraphics[width=.8\linewidth]{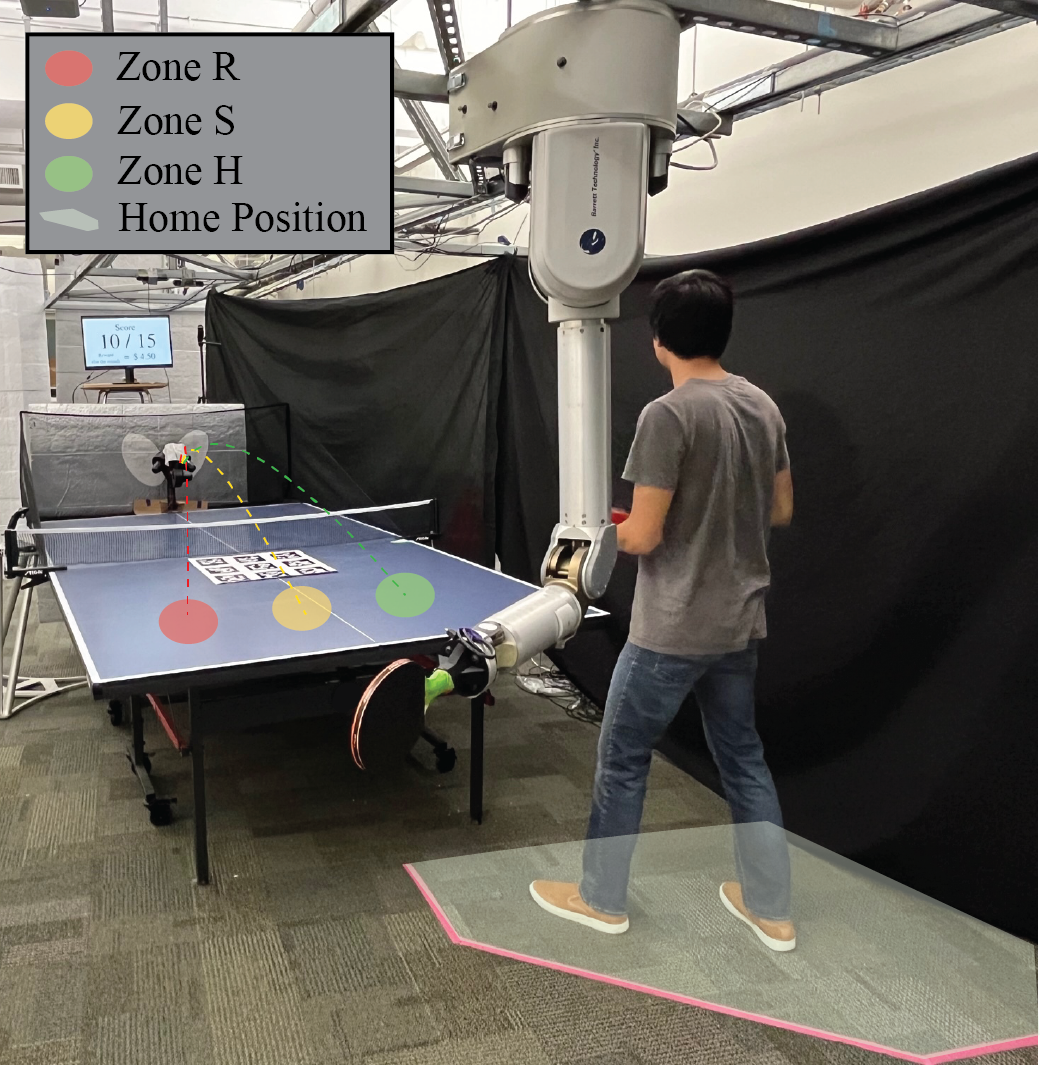}
     \vspace{-10pt}
     \caption{Setup of the collaborative table tennis system.}
     \label{fig:real_setup}
\end{figure}

\section{Related Work}

\noindent\textbf{Proximity in HRC --}
Cobots are envisioned to be in ``cage-free'' environments, sharing a workspace and collaborating closely with human workers \cite{Zaatari2019CobotPF}. 
As these cage-free systems can prove detrimental due to the potential for collision, several works have attempted to create safe human-robot collaboration via algorithmic advancements \cite{Peternel2017TowardsEC, DeLuca2006CollisionDA, Kuli2007PrecollisionSS}, designing different manipulation techniques to stop the robot from causing bodily harm in alignment with industry guidelines \cite{Romanov2019ARO}. 
\hl{Another approach has been to design compliant robots to ensure safety in proximate HRC settings~\cite{compliant_safecobot_2017, HaddadinAEH10, khatib_icra_04}.} 
Other work has looked into human-in-the-loop approaches that attempt to understand human intent for safe, proximate human-robot collaboration \cite{Gabler2017AGA}. However, while such techniques can produce safe behavior and address planning within a mixed-initiative teaming paradigm, these approaches do not address the perceived safety or trust of end-users, which are important factors in achieving high-performance HRC \cite{Busch2017LearningLM}.
Works in human factors have attempted to measure perceived safety through physiological signals that can indicate stress, fear, or anxiety \cite{POLLAK2020106469, fear_pupillary_dilation, dehais_physiological_2011}. Some of this work has been applied to industrial manipulators \cite{kulic_croft_2007, stress_around_robots}, though none have been applied in a fast-paced HRC setting to the best of our knowledge.

\begin{figure*}
     \centering
     \includegraphics[width=.8\linewidth]{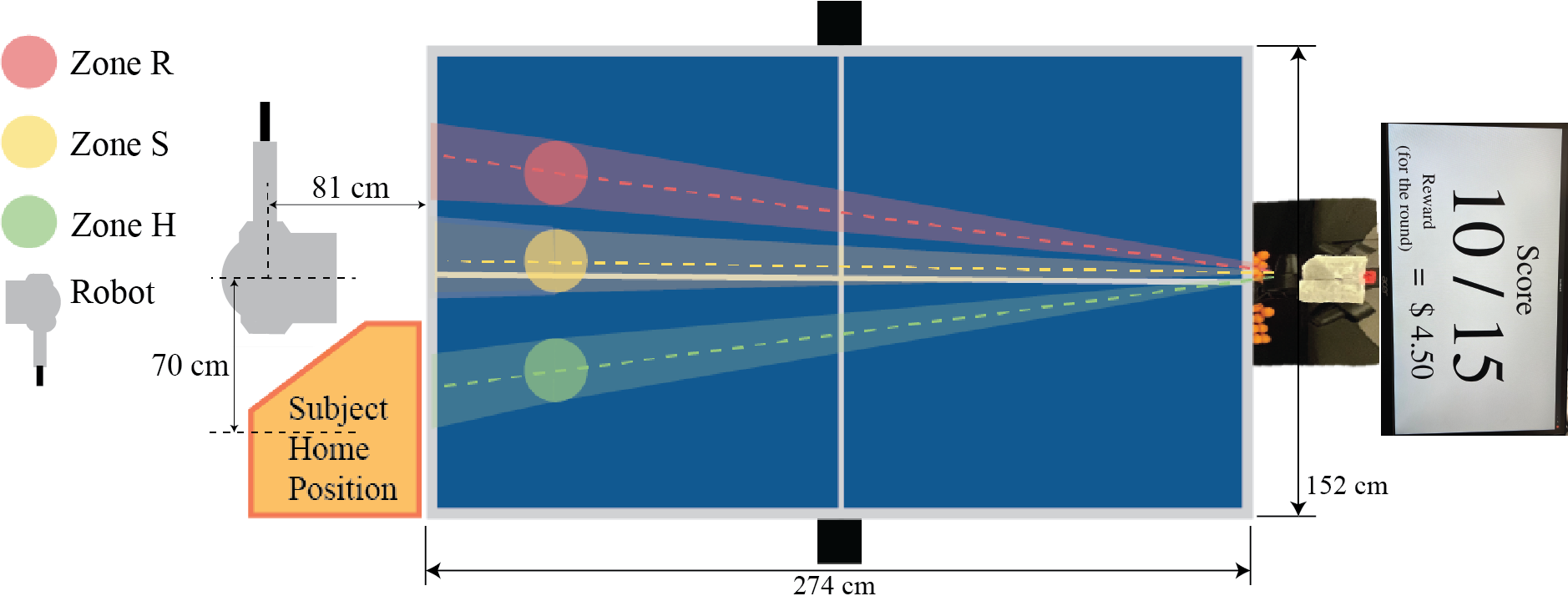}
     \vspace{-10pt}
     \caption{Top-down view of the task setup. Balls land in three zones: Robot (Zone R), Shared (Zone S), and Human (Zone H). The scoreboard on the right is updated in real-time.}
     \label{fig:topdown_setup}
\end{figure*}




\noindent \textbf{Communication in HRC --}
Prior work displayed that verbal communication in HRC affects collaboration as well as users' perception of the robot~\cite{Grigore2016TalkTM,MAVRIDIS201522,ROB-049}. Studies have shown that verbal communication between the robot and the user improves subjective and objective metrics on team performance and users prefer a speaking robot over a non-speaking one~\cite{Nikolaidis_verbalcommunication,clair_robotverbalfeedback,gaze_speech_collab,verbalfeedback_collab}. Furthermore, \cite{Delft2012EnhancingTP} found that autonomous agents who exchanged information regarding their intention (i.e., deliberative communication \cite{Butchibabu2016ImplicitCS}) collaborated with humans better than those sharing information in regard to the world state. It has also been suggested that a lack of truthfulness in a robot's statement can negatively affect a human's evaluation of a robot's abilities~\cite{Nikolaidis_verbalcommunication}.
Although these studies produced interesting findings about robot communication, the corresponding tasks did not have time pressure, deviating from real-time decision-making. 

In our experiment, the robot asserts its intention via a short verbal statement in a male/female voice to inform the participant that it will attempt to hit an incoming table tennis ball. The verbal communication is done in real-time by an agile robot conducting a high-speed striking maneuver. Deviating from prior literature, we additionally allow the human to communicate intent with the robot (bidirectional communication), creating a realistic, rapid, proximate HRC scenario where the human must reason about received communication, whether to communicate intent, which agent should hit the ball, and the ball trajectory, all within a short duration.


\noindent \textbf{Agile Robot Manipulation --}
Agile robotic systems that work alongside human beings is a long-standing goal in manufacturing, where high-speed robotic maneuvers are safely conducted with humans collaborating in proximity with the robot. In this way, manufacturing is able to take advantage of a human's ability to deal with uncertainty and variability while utilizing automation to its fullest \cite{Zaatari2019CobotPF}. However, prior work has noted that stress, anxiety, and fear levels of a human user around a fast-moving robot with the capability of causing physical harm are high~\cite{stress_around_robots,Salvini2022,Murashov2016-vi,saferobotmotionspeed_industrialrobots,Story2022,Butler2001,SHIBATA1998483}, presenting interesting challenges in producing high-performance HRC. Furthermore, as producing high-speed robotic maneuvers is a difficult challenge in itself, poor performance of the robot can adversely affect user trust~\cite{Hancock2011-bd, trust_in_time_crit, trust_after_mistakes, changing_reliability_trust}.
As such, to address these open problems in studying agile cobots, we study both the effects of varying skill level and communication on collaboration performance between humans and agile robot.

We utilize table tennis as a testbed for HRC, as this is a sport requiring fast reflexes and decision-making~\cite{tabletennis_timepressure}. In prior work, robotic systems have been developed to play singles table tennis~\cite{gomez2016using, buchler2022learning, Chen2020JointGA, Chen2020LearningFS, Chen2022FastLA, Ding2022LearningHS, Gao2020RoboticTT}. However, to the best of our knowledge, systems for doubles table tennis have not been explored. 



\section{Methodology}
\label{sec:study_design}


\subsection{Task Description}
\label{subsec:task_desc}

The participant's task is to team-up with a 7 degree-of-freedom Barrett WAM arm to return table tennis balls launched from a Butterfly Amicus ball launcher, as shown in \Cref{fig:topdown_setup,fig:real_setup}. To investigate the collaborative behavior between the human and the robot, we program the launcher to project balls to three regions: \zone{R} (the robot zone), \zone{S} (the shared zone), and \zone{H} (the human zone). The robot has the capability to return balls landing in \zone{R} and \zone{S}, and does not attempt to return the balls landing in \zone{H}. The human is free to return balls in any region, but those in \zone{R} are more difficult as those balls are farther from the subject's starting position (displayed in \Cref{fig:real_setup,fig:topdown_setup}). 

The task of returning the balls in \zone{S} is designed to be a \emph{shared task} to evaluate the collaboration between the human and the robot. In this region, both agents (robot and participant) can comfortably attempt the task. Note, we modify the standard doubles table tennis rules for our experiment such that the two players (i.e., the participant and the robot) do not have to alternate, and the human can decide which balls to attempt. The participants must start in the ``subject home position'' (shown as shaded region in \Cref{fig:topdown_setup,fig:real_setup}) before the ball is launched to ensure consistent starting configuration and difficulty to reach \zone{R}. 

The setup described meets the two criteria we set: 1) Low-reaction time in which the human has $\approx 0.6$ seconds to make a decision on whether to attempt to return a ball or yield it to the robot. 2) Shared Human-Robot Collaboration space following the definition of mixed-initiative teaming, where the interaction strategy is flexible, dynamic, and based on agents contributing to the task that they can do best~\cite{Allen1999MixedinitiativeI}. In our scenario, the subject is incentivized to rely on the robot for balls in \zone{R} due to the physical distance and share the \zone{S} with the robot. The participant score is displayed on a real-time scoreboard (Figure~\ref{fig:topdown_setup}); further details are in Section~\ref{subsec:study_proc}.

\subsection{Experiment Design}
\label{subsec:exp_design}
In our study, we seek to explore the impact and interplay of voice communication and robot skill level on team performance, human perceived safety, and human trust towards the robot. As such, within our collaborative task detailed in Section \ref{subsec:task_desc}, we design a study with three factors: (1) \emph{Communication}: enabled or disabled, (2) \emph{Robot Voice Sex}: male or female, (3) \emph{Robot Skill}: low and high. We propose a mixed-factorial study with a 2x2x2 design, with Communication and Robot Skill as within-subject factors and Robot Voice Sex as a between-subjects factor. While a complete within-subjects design is ideal for exposing participants to all conditions, we chose to investigate robot voice sex as a between-subjects factor to prevent participants from being aware that we are investigating sex/gender bias. We detail the three independent variables below. 

\paragraph{Communication:} Drawing inspiration from how volleyball players communicate deliberative intent, we enable the robot to vocally communicate assertive audio signals such as ``\emph{Mine}'' or ``\emph{Got it}'' through a speaker mounted on the robot arm. The utilization of short phrases is a requirement in agile robotics as intent must be quickly conveyed due to the low-duration task.  \hl{We define these Vocal, Assertive, INtention communication signals as ``VAIN'' signals for clarity and brevity on the form of communication used in our study. VAIN signals are provided in real-time as soon as a launched ball is detected by the vision system to indicate that the robot will attempt to return the ball. VAIN communication, when enabled, is used for all balls landing in both \zone{R} and \zone{S}. }

\paragraph{Robot Voice Sex:} We use a voice generation tool to create audio files for the phrases used in the communication factor with both a male and female tone to study whether/how the communication voice impacts the human-robot team's performance and the human's subjective perception of trust and safety. As noted, prior work has suggested that the sound of female vs.~male voice can have differing impacts on audibility, perceived assertiveness, reaction time, and quality~\cite{edworthy2003use,arrabito2009effects}. 

\paragraph{Robot Skill:} We have two levels of robot performance - low and high. The high performance model is  comprised of a well-tuned stroke primitive (detailed in \Cref{subsubsec:ball_strike_controller}) to return balls. The low performance model replaces the well-tuned primitive with a suboptimal primitive ($\sim$ 50\% of the time), which either misses the ball or does not successfully return the ball over the net. The success rates of the robot policies are summarized in \Cref{tab:robot_skill_calib}. 

Prior work has shown that robot skill has a positive correlation with trust on the robot \cite{Hancock2011-bd, trust_in_time_crit, trust_after_mistakes, changing_reliability_trust}. However, most findings have come from cognitive or physical tasks in virtual environments. Additionally, the impact of a robot's skill on perceived safety in HRC is under-explored. Exploring these topics in a setting with a physical, agile robot on a fast-paced task will lead to insights that can help in designing more effective cobots. 

Below, we detail our research questions under the following themes: Robot Skill and Deliberative Communication.
\subsubsection{Robot Skill} 
As skill is a critical factor in establishing effective collaboration, we are interested in answering the following questions pertaining to this factor through our study:
\begin{enumerate}[topsep=1ex,label=(\bf RQ\arabic*)]
    \item How does the robot's skill level impact team performance?  
    \item How does the robot's skill level impact the perceived safety of human collaborators?
    \item How does the robot's skill level impact human collaborators' trust on the robot?
    \item How does the robot's skill level impact human initiative in shared regions of the task?
\end{enumerate}

\subsubsection{VAIN Communication} We are interested in evaluating how Vocal, Assertive, INtention (VAIN) communication (i.e., a transparent robot that always announces its intention) during robotic table tennis motions affect the dynamics of human-robot collaboration. To this end we ask the following questions:
\begin{enumerate}[topsep=1ex,label=(\bf RQ\arabic*)]
    \setcounter{enumi}{4}
    \item Does VAIN communication help increase HRC team performance?
    \item Does VAIN communication increase perceived safety of the robot in HRC?
    \item Does communication increase people's trust on the robot in HRC?
    \item Does the sex of the robot's voice impact the perceived safety and trust of the robot?
    \item Does VAIN communication increase the robot's chance of taking initiative within the shared regions of task?
\end{enumerate}


\begin{figure*}
\centering
\begin{minipage}{.65\textwidth}
  \centering
  \includegraphics[width=\linewidth]{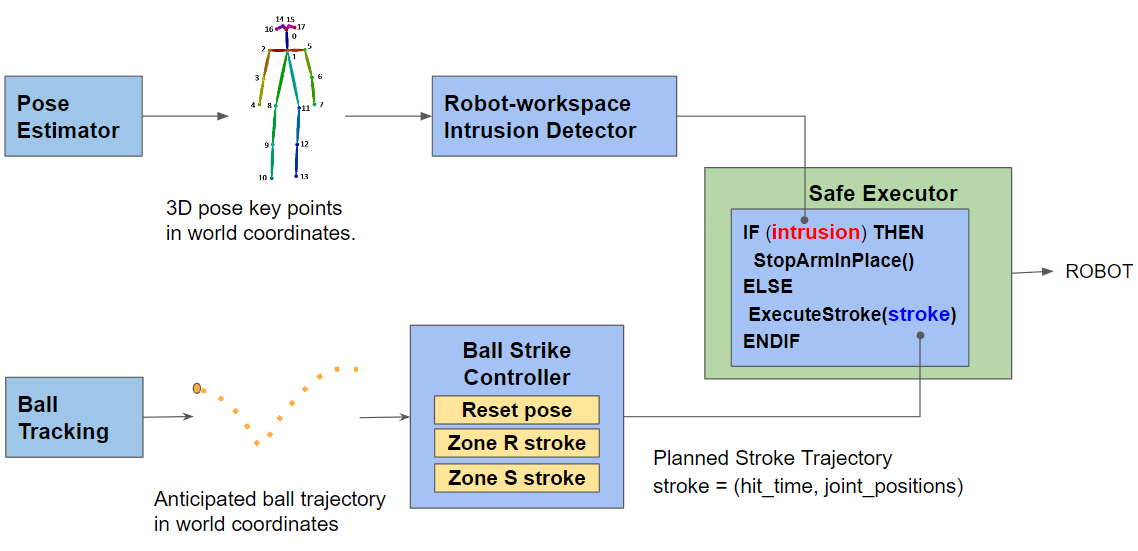}
    \caption{Overview of our collaborative table-tennis robot system.}
    \label{fig:sys_diag}
\end{minipage}%
\begin{minipage}{.34\textwidth}
  \centering
  \begin{subfigure}[b]{0.49\linewidth}
         \centering
         \includegraphics[height=4cm,width=\linewidth]{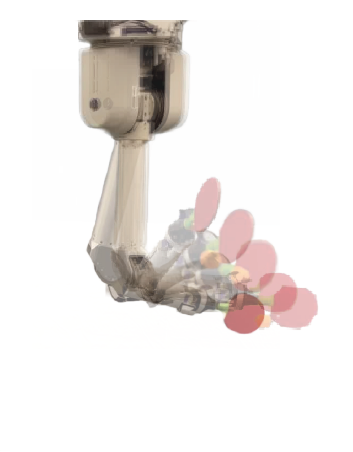}
         \caption{\zone{R} stroke}
     \end{subfigure}
         \begin{subfigure}[b]{0.49\linewidth}
         \centering
         \includegraphics[height=4cm,width=\linewidth]{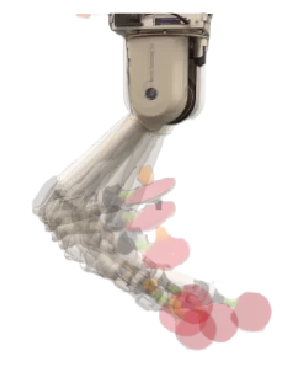}
         \caption{\zone{S} stroke}
     \end{subfigure}
    \caption{A visualization of the two robot stroke primitives that hits the balls that land in \zone{R} and \zone{S}.}
    \label{fig:viz_prims}
\end{minipage}
\end{figure*}

\begin{table}[ht]
    \centering
    \begin{tabular}{cccc}
    \hline
        {\bf Stroke Type} & {\bf Skill Level} & {\bf Hit Rate} & {\bf Success Rate}  \\ \hline
        \zone{R} & high & 94.36\% & 86.40\% \\
        \zone{R} & low & 93.66\% & 59.86\% \\ \hline
        \zone{S} & high & 86.54\% & 64.63\% \\
        \zone{S} & low & 83.89\% & 30.29\% \\
        \hline
    \end{tabular}
    \caption{Performance of robot strokes on balls that land in \zone{R} and \zone{S} for the different skill levels. ``Hit'' denotes the robot successfully hit the ball with the paddle. ``Success'' indicates the ball is successfully returned over the net.}
    \label{tab:robot_skill_calib}
    \vspace{-15pt}
\end{table}

\subsection{System Design}
\label{subsec:sys_design}
Our robotic system has four main components to achieve various levels of robot skills and ensure safety, as shown in \Cref{fig:sys_diag}. First, Ball Detection enables the robotic system to ``see'' and predict the ball trajectory in order to decide when to hit. Second, if the robot is able to strike the incoming ball (i.e., the ball must be in \zone{R} or \zone{S}), the Ball Strike Controller utilizes ProMP~\cite{gomez2016using} policies to execute the corresponding pre-trained strikes for the zone. Third, Human Pose Estimation detects the 3D pose of the human. Lastly, Safe Execution determines if any human pose keypoints have entered the robot safety region, which leads to an immediate termination of the robot trajectory to avoid any possible collision between the robot and the human. 


\subsubsection{Ball Tracking}
To accurately detect the table tennis balls, we utilize a multi-camera vision system consisting of three Stereolabs ZED2 cameras. Classical computer vision techniques, including background subtraction and color thresholding (from OpenCV \cite{opencv_library}), are applied to identify the ball pixels within the image. With this, an estimate of the ball position in the world frame is calculated based on each camera's relative position to an AprilTag \cite{wang2016apriltag} on the ping pong table. All three ball poses are merged by an Extended Kalman Filter (EKF) \cite{kalman_filter,julier1997new} to produce an accurate fused estimate of the ball. A dynamics model that factors in drag, gravity, and the coefficient of restitution of bounce is used to rollout the EKF's expected future ball trajectory, which is then utilized by the Ball Strike Controller to determine the right time to execute a stroke. 

\subsubsection{Ball Strike Controller}
\label{subsubsec:ball_strike_controller}
We trained two probabilistic movement primitives (ProMPs)~\cite{gomez2016using} -- one for \zone{R} and \zone{S} -- using ten kinesthetic demonstrations of successful returns for each. We model the joint position trajectories of the stroke as a weighted activation over 50 radial basis functions in the phase space $[0, 1]$ and learn the weight distribution that maximizes the likelihood of the demonstrated trajectories through the Expectation-Maximization (EM) procedure outlined in~\citet{gomez2016using}. The trained stroke primitives are visualized in \Cref{fig:viz_prims}. While the learned primitive weights can be conditioned to pass through waypoints based on the ball trajectories, we execute the unconditioned mean of the learned primitive due to the possible unsafe waypoints of a conditioned ProMP. Doing so effectively ensures that the stroke performed by the robot for each zone is pre-determined, thereby keeping the robot's behavior predictable and safe for the purposes of the study.


During our experimentation, the sequence of ball launch locations is predetermined, providing the robot with access to information about which region each ball will land. To avoid the participant from obtaining information about incoming ball trajectories, for each strike, the robot starts at a common reset pose and moves to the initial pose of the primitive only once a ball is detected by the Ball Tracking component.
The common reset pose was determined to be a weighted average of the initial poses of the \zone{R} and \zone{S} strokes. 
Once a ball is detected, the system utilizes the anticipated ball trajectory to calculate the time of impact and the approximate paddle position at the time of impact, which is used to trigger an agile robot table tennis return.

The stroke parameters, such as duration and hit phase were tuned to return as many balls in each zone as possible. These strokes serve as the \emph{high} skill model of the robot. In order to create a \emph{low} skill model of the robot, we introduce mistiming to the strokes ~50\% of the time. This offsetting during the hit phase causes the robot to poorly return the ball or completely miss. We report the performance of the robot using these two skill models over all the ball sequences considered in the experiment trial in \Cref{tab:robot_skill_calib}. 


\subsubsection{Human Pose Estimation}
We utilize a ZED2 camera for tracking human pose. The ZED software development kit provides access to a body tracking module that outputs skeleton-based human keypoints. These keypoints indicate the locations of major joints within the human body such as the head, shoulders, elbows, hands, knees, and feet. We use the 18-point 3D pose format (OpenPose~\cite{8765346} pose format) in our system to track the pose of participants. Each 3D keypoint consists of an X, Y, and Z position of a human joint in the camera frame. These keypoints are then transformed to world coordinates relative to an AprilTag \cite{wang2016apriltag} on the table, which is used in the safe execution module to avoid the robot colliding with humans.

\subsubsection{Safe Execution}
All control requests to the robot are routed through the safe execution module as shown in \Cref{fig:sys_diag}. A fixed 3D unsafe region is created around the robot based on the reachability range of the robot strokes calculated by the Ball Strike Controller. The skeleton keypoints from the Human Pose Estimation module are passed as input into the safe execution module to detect if any part of the participant's body enters the unsafe region. If so, any robot motion in progress is immediately terminated by sending a stop command directly to the low-level controller and all further control requests are rejected. Once keypoints are no longer detected within the unsafe region (i.e., the human is outside of the robot striking area), control requests are accepted again and the robot is reset to its default joint positions. Additionally, the maximum end-effector speed of any stroke is limited to 3.5 m/s to ensure safety even if the unlikely scenario of a collision were to occur.

\subsection{Metrics}
\label{subsec:metrics}

\subsubsection{Objective Metrics}
\hl{The team performance metric is the sum of successful returns both the robot and participant achieve together in a randomized sequence of 30 balls. This randomized sequence consists of 8 balls in \zone{H}, 14 in \zone{S}, and 8 in \zone{R}}. In order to control the impact of the participants' prior table tennis skill to the overall performance score, we measure the participant's skill level in performing the described task in a calibration phase where the participant returns a sequence of 30 balls on their own. We record for each ball, the zone in which it landed, whether it was hit, and if the ball was returned successfully. Similarly, in the main experiment, we record for each ball the landing zone, who attempted to hit it, whether it was hit, and if the return was successful. 

As an additional dependent variable, we create a bidirectional communication channel by allowing the subject to speak the phrases (``Mine'', ``Got it'') to convey their intent of returning the ball. Here, we observe if robot skill level or robot communication incentivized the human to communicate their intent. Note we explicitly state that participant communication towards the robot is optional and not required for the robot to stop its stroke. The robot always yields given participant communication, responding with either ``\emph{Okay}'' or ``\emph{Alright}''. Due to the latency of speech recognition and the limited timing allowed in our task, we choose a Wizard-of-Oz approach \cite{dahlback1993wizard} to provide a timely response to the participant's declaration of intent - a button is pressed by the experimenter to trigger immediate stopping of the robot movement and provide voice responses. 

\subsubsection{Surveys}
We conduct a pre-experiment questionnaire that includes self-reported demographics - age, gender, and dominant hand in playing table-tennis, mini-IPIP\cite{donnellan_mini-ipip} to measure the big-five personality factors, and negative attitude towards robots~(NARS)\cite{syrdal_negative_nodate} to measure the subject's predisposition towards robot interactions. Each of these serve as potential confounds within our analysis.
After each within-subject condition (described in depth within Section \ref{subsec:study_proc}), the subject is asked to fill out a questionnaire assessing their interaction with the robot. This questionnaire includes the Godspeed questionnaire \cite{bartneck2009measurement}, measuring anthropomorphism, likeability, perceived intelligence, and perceived safety, and includes the trust in automated systems questionnaire \cite{jian2000foundations} to measure perceived trust.

\subsection{Study Procedure}
\label{subsec:study_proc}
To assign participants across all conditions uniformly, we sample sequences in blocks of four samples with an even split of robot voice sex in a pseudo-random fashion. The sequence order of conditions is fully randomized. The same seed is used for consecutive blocks of samples where the last 11 samples are manually assigned to ensure the gender of participants are balanced across robot voice sex. 

For each participant, we start by obtaining informed consent. In the consent form, we note that our experiment may have deception approved by our IRB protocol. This hidden deception helps incentivize participants to be more involved and competitive in the task by deducting 10 cents from the study compensation for each ball missed irrespective of whether it was the human or the robot. We further utilize a digital scoreboard showing the current score alongside the loss in compensation in the current experiment block (\Cref{fig:topdown_setup}) to emphasize the importance of human-robot collaboration. Note, participants were debriefed about the deception at the end of our study and were paid the full amount of compensation. 

\begin{figure*}
    \centering
    \begin{subfigure}[b]{0.3\linewidth}
        \centering
        \includegraphics[width=\linewidth]{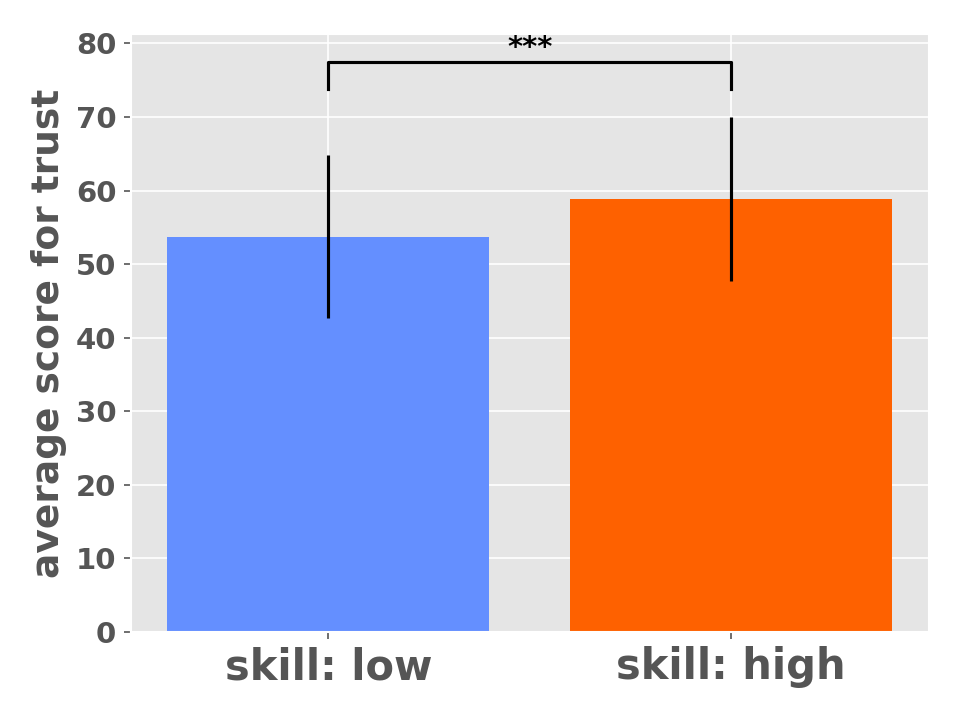}
        \vspace{-10pt}
        \caption{Effect of Robot Skill on the perceived \\ trust of the system.}
        \vspace{-5pt}
        \label{subfig:skill_trust}
    \end{subfigure}
    \begin{subfigure}[b]{0.3\linewidth}
        \centering
        \includegraphics[width=\linewidth]{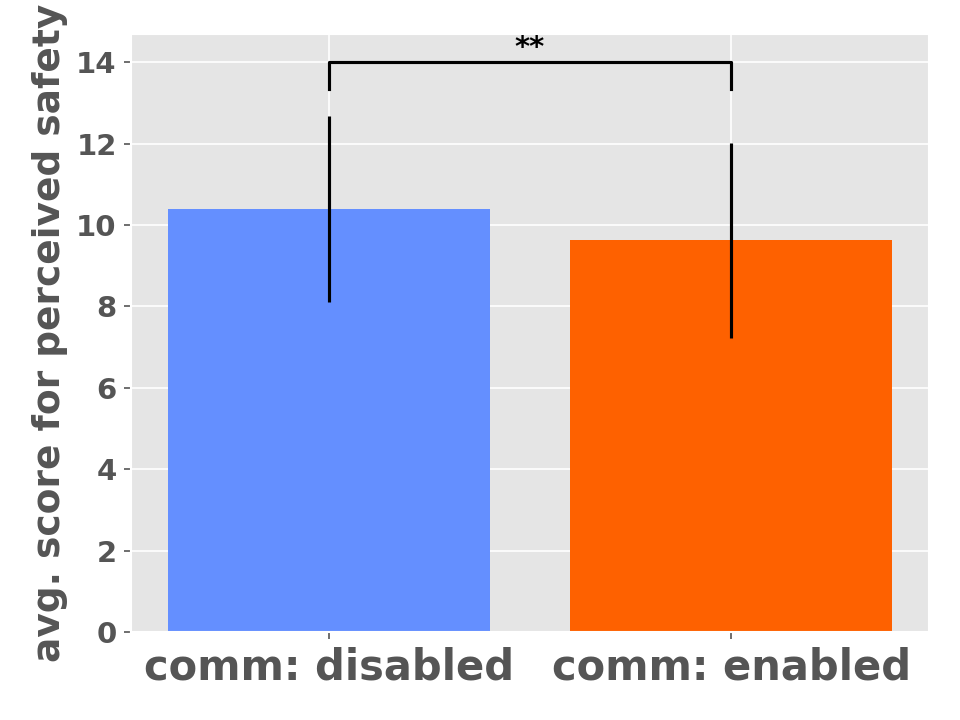}
        \vspace{-10pt}
        \caption{Impact of VAIN Communication on \\ the perceived safety of the system.}
        \vspace{-5pt}
        \label{subfig:comm_safety}
    \end{subfigure}
    \begin{subfigure}[b]{0.3\linewidth}
        \centering
        \includegraphics[width=\linewidth]{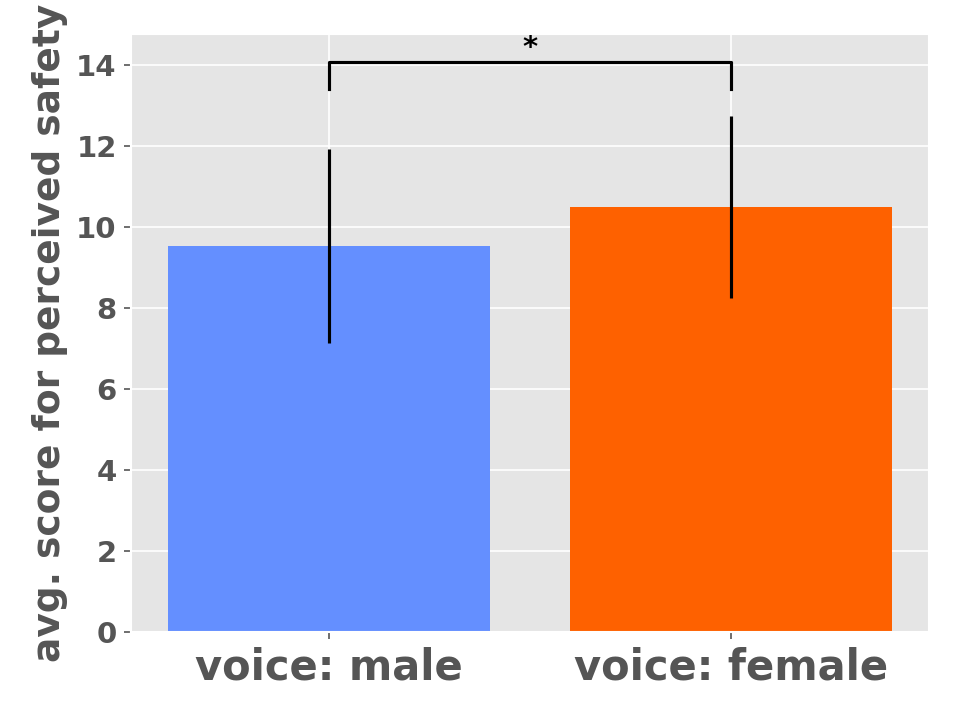}
        \vspace{-10pt}
        \caption{Effect of robot voice sex on the perceived safety of the system.}
        \vspace{-5pt}
        \label{subfig:voice_safety}
    \end{subfigure}
    \caption{Effect of robot skill, VAIN communication and robot voice on the perceived trust and safety.}
    \vspace{-5pt}
    \label{fig:iv_safety}
\end{figure*}

Upon completion of informed consent, we give the participant a pre-experiment survey to assess self-reported demographics, personality factors, and negative attitude towards robots. Once participants have completed the survey, we perform skill calibration for a sequence of 30 balls to access the subject's performance on the task without the robot. The subject is allowed a two-minute warm up period before the calibration task begins. After the calibration phase, we move into the main study. We begin by providing a description of the collaborative study and show a demonstration over a sequence of 6 balls with the high-skill model of the robot. The demonstration provides the participant with an idea of the robot capability and show that it is safe to step into the robot's configuration space to intercept a ball. We additionally demonstrate the high-skill model of the robot with communication enabled and show how both the robot and human can declare intent, and we inform the participant that communicating with the robot is optional. The participant is allowed a warm-up phase, actively collaborating with the robot under no communication. This allows the participant to get used to the robot's presence and minimize the learning effect in the first sequence. During this warm-up, the ball sequence order is fixed and evenly distributed across all zones. 

For each condition of the 2x2 within-subject block design, we describe whether the robot has communication enabled and stress that the performance is independent of previous trials. Information about robot performance is kept secret from the participant. Each participant will experience a total of four trials varying across Robot Skill and Communication. The human-robot team will play a 30 ball sequence for each condition, after which the subject is asked to evaluate the interaction with the Godspeed~\cite{bartneck2009measurement} and Trust in Automated Systems questionnaires~\cite{jian2000foundations}. Upon completion of the four trials, we debrief the subject about the experiment, detail the deception/concealment, and obtain consent to use their data.

\section{Results}
We recruited 46 participants for this in-person IRB-approved study. Each participant was compensated \$20 for completing our study. Four participants were excluded from our analysis as we experienced issues with the cable tension of the robot arm, resulting in visibly degraded robot performance during the experiments. In our results, we report on 42 participants (sex: 20 Male, 20 Female, 2 Prefer not to say) and age ($M = 25.43$, $SD = 3.5$).

For each measure of trust, perceived safety, team performance, and anthropomorphism, we create a mixed-effects multiple regression model with metrics of interest as the dependent variable. The independent variables are skill, communication, and robot-voice, and we include covariates such as gender, age, personality factors, calibration score, etc. We then applied an Analysis of Variance (ANOVA) on each model to answer our research questions. The mixed-effects model accounts for the repeated measures nature of each subject completing the 2x2 within-subject blocks. For each model, we report the transforms used, results from the Shapiro-Wilk test for normality, and the Levene's test for homoscedasticity for the independent variables in the Appendix. We measure significance as $\alpha < 0.05$. 
If the assumptions for normality or homoscedasticity do not hold, we report a non-parametric assessment using Friedman's test followed by an all-pairs Nemenyi test comparing the within-subject factors of robot-skill and communication, and do not consider the influence of other covariates.

\subsection{Perceived Trust}
Across all conditions of gendered voice, skill, and VAIN communication, we compared how participants rated their levels of trust after completing each block comprising a sequence of 30 balls. We find that when the robot performed poorly, the trust in the robot reduced significantly \checkstat{($F(1,124)=18.687$, $p<.001$)}, as shown in \Cref{subfig:skill_trust}. This finding concurs with many previous human-robot interaction studies evaluating trust on the robot with varying skill levels~\cite{Hancock2011-bd} and addresses \RQ{RQ3}. However, we find that VAIN communication did not impact the level of trust on the robot \checkstat{($F(1,124)=0.463$, $p=.497$)}, concluding for \RQ{RQ7} that communication does not improve trust on the robot.


\subsection{Perceived Safety}
 To investigate \RQ{RQ6}, we analyze the relationship between VAIN communication and perceived safety. We observe that this communication significantly decreased perceived safety \checkstat{($F(1,124)=7.070$, $p=.009$)}, as shown in \Cref{subfig:comm_safety}. We hypothesize that the constant announcement of the robot's intention induces more anxiety and agitation, which leads to lower scores within the perceived safety sub-scale. Additionally, we observe that the sex of the robot voice was a significant factor affecting participants' perception of the robot. On average, the robot equipped with a female voice was perceived safer than the robot with a male voice \checkstat{($F(1,38)=7.079$, $p=.011$)}, as shown in \Cref{subfig:voice_safety}, addressing \RQ{RQ8}. The participant's gender also influences the perceived safety score, with male participants rating the robot as safer than female participants \checkstat{($F(2,38)=4.709$, $p=.015$)}. In addressing \RQ{RQ2}, we find that the robot skill level is marginally significant for perceived safety, with a robot with a higher skill level perceived to be safer \checkstat{($F(1,124)=2.762$, $p=.099$)}.



\subsection{Team Performance}
\begin{figure}
    \centering
    \begin{subfigure}[t]{0.49\linewidth}
        \includegraphics[width=1.1\linewidth]{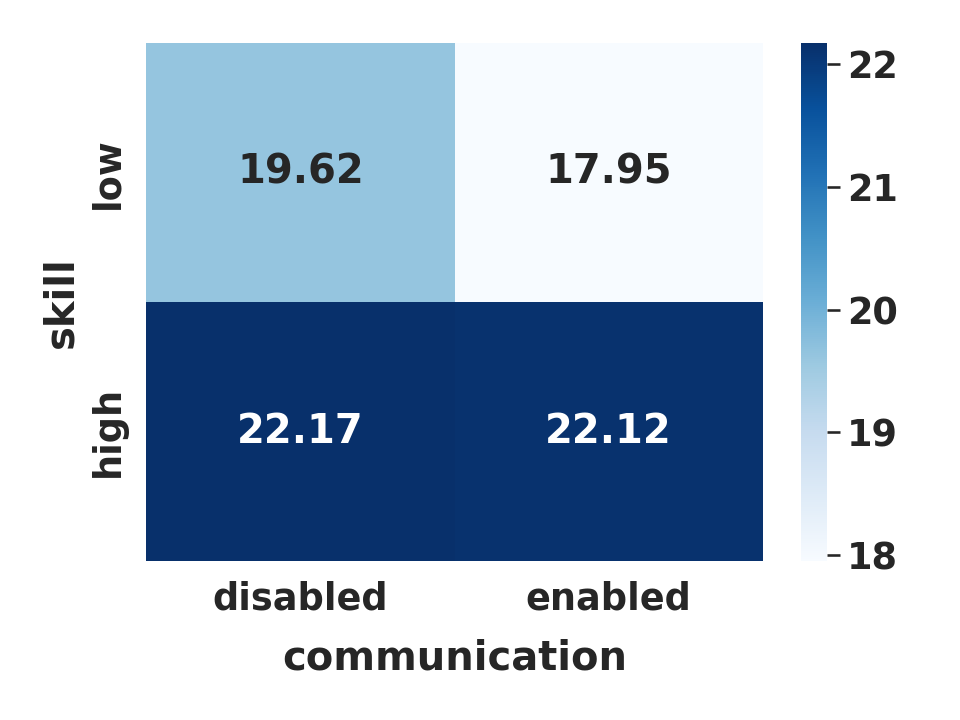}
        \caption{\hl{Average team score out of the 30 possible points.}}
        \label{subfig:comm_skill_team_perf}
    \end{subfigure}
    \hfill
    \begin{subfigure}[t]{0.49\linewidth}
        \centering
        \includegraphics[width=1.1\linewidth]{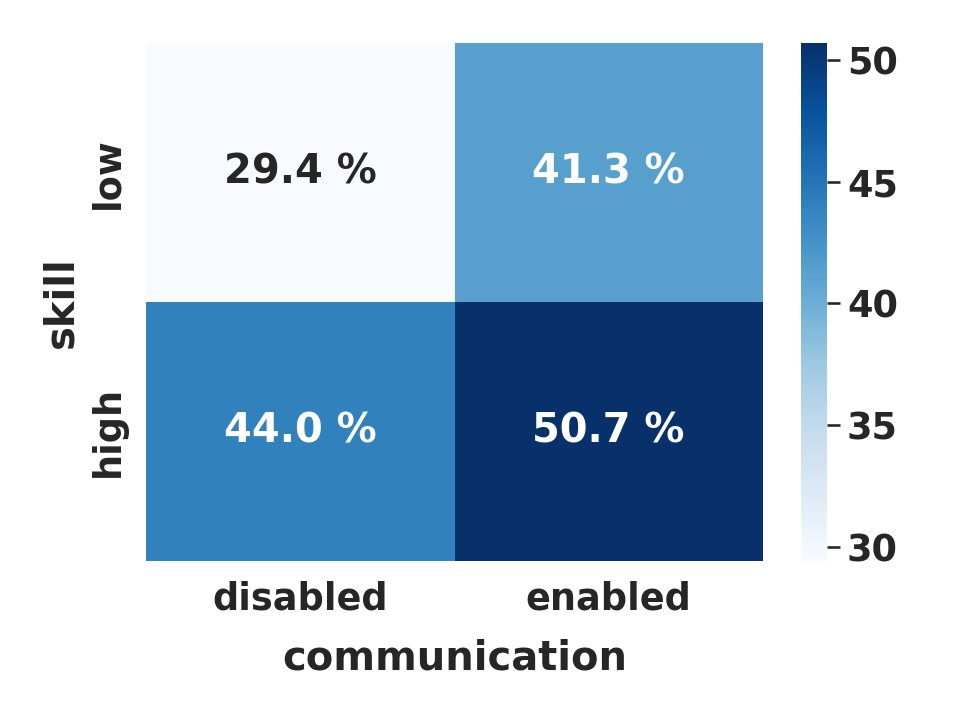}
        \caption{\hl{Percentage of shared tasks yielded to the robot on average.}}
        \vspace{-5pt}
        \label{subfig:comm_skill_shared_yielded}
    \end{subfigure}
   \caption{Effect of robot skill and VAIN communication on the team performance.}
    \vspace{-5pt}
    \label{fig:perf_skill_comm}
\end{figure}

For \RQ{RQ1}, we find that a higher skill level of the robot significantly improves the team performance \checkstat{($F(1,123)=69.25$, $p < .001$).} Additionally, the skill of the human, measured by their score on the calibration task, improved team performance \checkstat{($F(1,34)=29.929$, $p<.001$)}. The gender of the participant also significantly impacts team performance, with male participants having higher team performance on the  task \checkstat{($F(2,34)=6.314$, $p = 0.005$)}. In addressing \RQ{RQ5}, we find that VAIN communication degrades HRC team-performance \checkstat{($F(1,123)=4.514$, $p = .037$)}. We find an interaction effect between communication and skill also being significant \checkstat{($F(1,123)=4.0266$, $p = 0.047$)}, indicating that communication significantly hurts team-performance when the robot's skill is low, and affects team performance minimally when robot skill is high, as visualized in \Cref{subfig:comm_skill_team_perf}.

To answer \RQ{RQ4} and \RQ{RQ9}, we perform a non-parametric all pairs Nemenyi test, as some independent variables failed the Levene's test. The Friedman's test reported that the condition: skill + communication significantly impacts the number of shared tasks attempted by the robot. This metric indicates that the participant yielded these tasks to the robot ($\chi^2(3)=18.071$, $p<.001$). Comparing the different conditions with an all pairs Nemenyi test, we find that when the communication is disabled, robot with higher skill was given more chances to play the shared task ($p=.034$). \hl{For \RQ{RQ9} we find that communication improves the robot's odds of playing the shared task, as depicted in \Cref{subfig:comm_skill_shared_yielded}, but the difference is not statistically significant.}

Our setup additionally allowed the participant to declare their intent to the robot verbally, similar to what the robot does in the VAIN communication condition. Participants rarely made use of this channel to declare their intent, with a median of 3 declarations in a sequence of 30 balls, with 45.2\% of the participants not even using the channel. We note that this metric did not correlate with other factors such as calibration score, personality factors, etc.


\subsection{Anthropomorphism}
We observe that the robot with high skill was rated higher in terms of anthropomorphism \checkstat{($F(1,120)=4.762$, $p = .031$)}. We hypothesize this is due to the fluid nature and legible behavior of successful strokes compared to the jerky movement exhibited when the robot fails to return balls. Furthermore, we find VAIN communication was marginally significant in impacting perceived anthropomorphism \checkstat{($F(1,120)=3.047$, $p = .083$)}. A participant's conscientiousness score (i.e., their propensity towards organization and attention to detail) positively impacted how much they anthropomorphized the robot \checkstat{($F(1,35)=9.257$, $p=.004$)}. Conversely, a participant's rating on the NARS subscale negatively influenced the participants perception of anthropomorphism \checkstat{($F(1,35)=7.084$, $p=.012$)}. We also see that participants that are detail-oriented found the robot more anthropomorphic, while participants that are worried about robots having emotions found the robot less anthropomorphic.

When communication is enabled, the male/female voice of the robot influences the perception of anthropomorphism. The robot using a female voice is rated as more anthropomorphic than the robot with the male voice \checkstat{($F(1,120)=4.762$, $p=.031$)}. The gender of the participant and its interaction with the robot voice were not significant for anthropomorphism.

\subsection{Likeability}
We analyze the impact of the skill and the communication factor on this dependent variable through non-parametric tests as this variable failed to meet the assumption of normality. The Friedman's test indicated that the condition: skill+communication significantly impacts likeability \checkstat{($\chi^2(3)=21.171$, $p<.001$).} To further compare the different conditions, we perform an all-pairs Nemenyi test and find that when communication is disabled, higher robot skill increases likeability \checkstat{($p=.043$)}, and with low robot skill, communication tends towards significance in improving likeability \checkstat{($p=.075$).}

\section{Discussion}

Existing work has shown that verbal communication increases team performance, and that users prefer robots with vocal capabilities. Further, it has been shown that cues which signal specific robot motions (e.g., emitting a sound for a particular motion or non-verbal cues, such as eye contact, for handover tasks) improve perceived safety~\cite{stress_around_robots}. However, our results show that in a dynamic, mixed-initiative setting with proximate collaboration, VAIN communication negatively affects team performance and perceived safety. We hypothesize that VAIN communication overloads the subject's mental resources given the short timeframe ($\sim$ 0.6 seconds) for task execution~\cite{information_overload}. Within this time, participants must reason about the ball trajectory to determine who should return the ball, process received communication, and optionally communicate information. Extra information may make planning more strenuous, resulting in lower team scores on average.

Moreover, VAIN communication, while transparent, provides a signal to the human just before the robot makes a sudden and fast movement. \hl{Such explicit signals could convey to the user that the robot could be capable of overriding the person to take an unsafe action, contributing further to the ambience of physical danger and a decrease in perceived safety. Based on our current findings, we posit that VAIN communication should not be used in fast-paced human-robot teaming systems; instead, robots should be primarily defensive and avoid the appearance of conflicting with a human's autonomy.} However, prior work \cite{Zhang2021ExploringCR} studying sound caused by robots found that louder robots are described as ``scary'' and ``dangerous,'' \hl{indicating that other sound profiles of VAIN communication could potentially lead to improvement}. 

\hl{Additionally, we found that VAIN communication from human to the robot was rarely utilized, suggesting that participants felt more comfortable communicating to the robot implicitly with their body motion as opposed to explicitly declaring their intent with voice in fast-paced tasks. While providing general feedback, some participants noted that talking to the robot was redundant, as intercepting the ball causes the robot to stop regardless. As such, we recommend that agile collaborative systems be developed that can infer human intent from physical motion rather than relying on explicit communication. Nonetheless, it is possible other forms of voice framework and design beside VAIN communication could show benefits to using voice communication in agile collaboration. Humans in agile team sports (volleyball, tennis doubles, etc.) all incorporate some form of VAIN communication to avoid conflict, so further exploration is needed to understand how to achieve effective communication with robots. }


\textbf{Limitations --} As indicated in Table \ref{tab:robot_skill_calib}, the success rate of the high robot skill level is not perfect. The number of failures vary across each trial, where the timing of these failures can alter the subject’s perception of the robot’s skill. While not common, some trials had a series of missed swings at the beginning of a sequence on a robot with high skill. This can lead the subject to believe the robot is poor performing, and thus receiving fewer attempts at the ball for the rest of the sequence. Additionally, our participants were largely students recruited from a university campus. \hl{The use of deception in the study also creates a high-stakes environment which can create more pronounced effects on our findings compared to a lax environment; however, we believe that our current design more accurately represents real-world HRC scenarios where failure results in negative consequences (e.g., a physical collision).}

\textbf{Future Work --} In future work, we would like to conduct an extension to this study to identify agile robot system characteristics that lead to increased perceived safety, perceived trust, and HRC performance. Our system was designed to conservatively avoid any collision with the participants, and we plan to improve the system with tighter safety boundaries to allow closer interactions. 

\vspace{-1mm}

\section{Conclusion}
In this work, we utilize a modified human-robot teaming, table-tennis doubles task in a human-subject experiment to gain insights that can inform the design of agile, collaborative robots. Specifically, we study a mixed-initiative teaming scenario with quick decision-making from humans and robots in a shared collaboration space where either agent can successfully accomplish the task. We investigate how robot voice sex, robot skill level, and deliberative communication impact perceived safety and trust in agile, proximate human-robot collaboration.
We conduct a human-subjects experiment with 42 participants and find robot skill is positively correlated with trust, extending skill-trust assessment in prior studies to agile, proximate HRC. Furthermore, interestingly, we find that VAIN communication degrades human-robot teaming performance and decreases perceived safety, indicating the need for more research enabling safe and trustworthy agile robots that operate in proximity with humans.

\begin{acks}
This research was supported by a gift award by Google to the Georgia Tech Research Foundation.
\end{acks}

\bibliographystyle{ACM-Reference-Format}
\balance
\bibliography{references}

\end{document}


\appendix
\section{Analysis Summary}

\begin{table*}[!ht]
  \renewcommand{\arraystretch}{1.5}
  \centering
  \caption{\label{table:model_assumptions} This table details the dependent variables, independent variables, and covariates included in each model. Additionally, if the statistical test was a parametric test, the tests for normality and homoscedasticity assumptions are included. Any transforms used on the dependent variable are also included. }
    \begin{tabular}{ | c | c | c |  c | c |c |}
      \hline
      \textbf{DV} & \textbf{Transform}         & \textbf{IV/Covariates}         & \textbf{Statistical Test}   & \textbf{Assumptions} & \textbf{p-values} \\
      \hline
        Trust        & boxcox   &
        Skill, Human Gender, Communication & rANOVA & Shapiro-Wilk & $p=.068$  \\ \cline{5-6}

        & & & & Levene's (skill) & $p=.950$ \\ \cline{5-6}

         & & & & Levene's (comm) & $p=.860$ \\ \cline{5-6}

          & & & & Levene's (voice) & $p=.971$ \\ \cline{5-6}

           \hline
        Perceived Safety        & N/A   &
        Skill, Human Gender,  & rANOVA & Shapiro-Wilk & $p=.432$  \\ \cline{5-6}

        & & Communication, Robot Voice & & Levene's (skill) & $p=.719$ \\ \cline{5-6}

         & & & & Levene's (comm) & $p=.454$ \\ \cline{5-6}

          & & & & Levene's (voice) & $p=.255$ \\ \cline{5-6}

                     \hline
        Perceived Safety        & N/A   &
        Skill*Communication, Human Gender,  & rANOVA & Shapiro-Wilk & $p=.761$  \\ \cline{5-6}

        & & Robot Voice, Human Skill, Handedness,  &  & Levene's (skill) & $p=.299$ \\ \cline{5-6}

         & & Team Sport Experience, NARS Emotion Subscale &  & Levene's (comm) & $p=.705$ \\ \cline{5-6}

          & & & & Levene's (voice) & $p=.818$ \\ \cline{5-6}

                              \hline
        Anthropomorphism        & N/A   &
        Skill*Communication*Robot Voice,  & rANOVA & Shapiro-Wilk & $p=.379$  \\ \cline{5-6}

        & & Human Gender, Handedness,  &  & Levene's (skill) & $p=.516$ \\ \cline{5-6}

         & & Conscientiousness, NARS Emotion Subscale &  & Levene's (comm) & $p=.072$ \\ \cline{5-6}

          & & & & Levene's (voice) & $p=.625$ \\ \cline{5-6} 

                                   \hline
        Likeability        & N/A   &
        Skill+Communication Condition  & Friedman's & Shapiro-Wilk & $p<.05$  \\ \cline{5-6}

        & &  &  & Levene's (skill) & $p=.744$ \\ \cline{5-6}

         & & &  & Levene's (comm) & $p=.065$ \\ \cline{5-6}

          & & & & Levene's (voice) & $p=.917$ \\ \cline{5-6} 

          \hline
                  Number Shared Task       & N/A   &
        Skill+Communication Condition  & Friedman's & Shapiro-Wilk & N/A  \\ \cline{5-6}

        Yielded to Robot & &  &  & Levene's (skill) & $p=.146$ \\ \cline{5-6}

         & & &  & Levene's (comm) & $p<.05$ \\ \cline{5-6}

          & & & & Levene's (voice) & $p=.508$ \\ \cline{5-6}

    \hline

    \end{tabular}
    \end{table*}

We considered the following covariates along with the independent variables of skill, comm, group and it's interaction effects, and were included in the model based on it's significance or if it lowered the AIC of the model.
\begin{enumerate}
    \item participant gender
    \item calibration score
    \item participant age
    \item dominant hand for play
    \item familiarity with AI or robotics
    \item experience with team sports
    \item big-five personality factors from miniIPIP
    \item subscale factors of NARS
\end{enumerate}

\newpage
\section{Participant self-reported demographics}

\begin{table*}[ht]
\renewcommand{\arraystretch}{2.5}
\begin{tabular}{|m{4cm}|m{7cm}|} \hline
Number of participants & 42 \\ \hline
Participant Sex & Male: 20, Female: 20, Prefer not to say: 2 \\[0.6em]\hline
&\\[-3em]
{Age} & {\centering\includegraphics[scale=0.39]{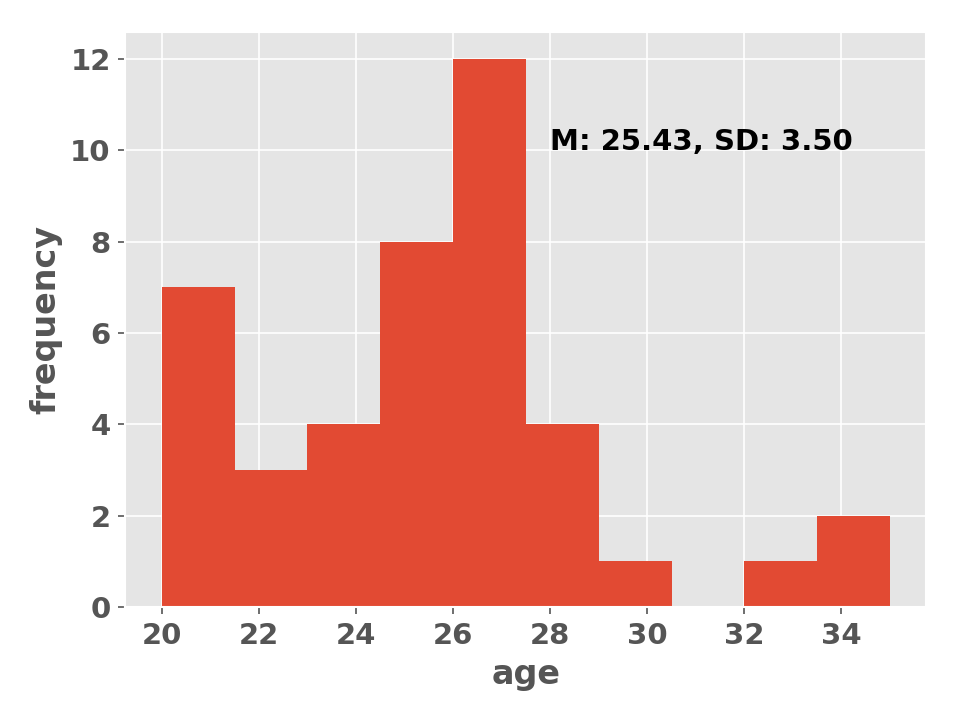}}\\ \hline
Handedness & Right: 40, Left: 2 \\\hline
&\\[-2em]
Highest Education & $\vcenter{\begin{itemize}[leftmargin=*] \item High school graduate: 1 \item Some college: 9 \item 4 year degree: 18 \item Professional degree: 14 \end{itemize}}$\\
&\\[-2em]\hline
&\\[-3em]
{Familiarity with AI \& Robotics (scale 1 to 5)} & {\centering\includegraphics[scale=0.39]{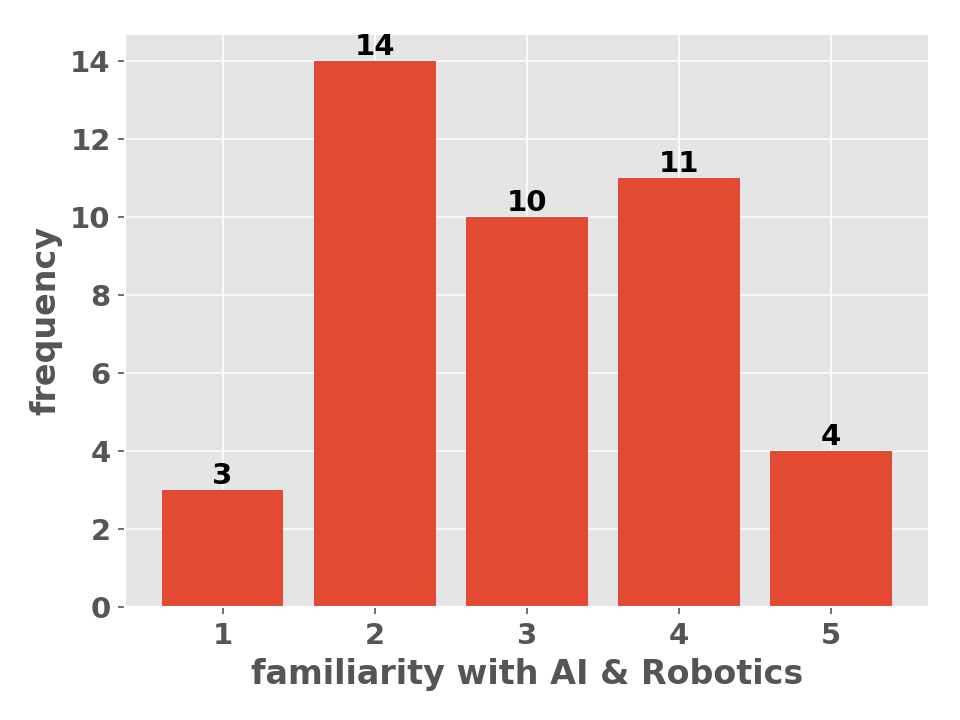}} \\[0.6em] \hline
&\\[-3em]
{Experience with team sports (scale 1 to 5)} & {\centering\includegraphics[scale=0.39]{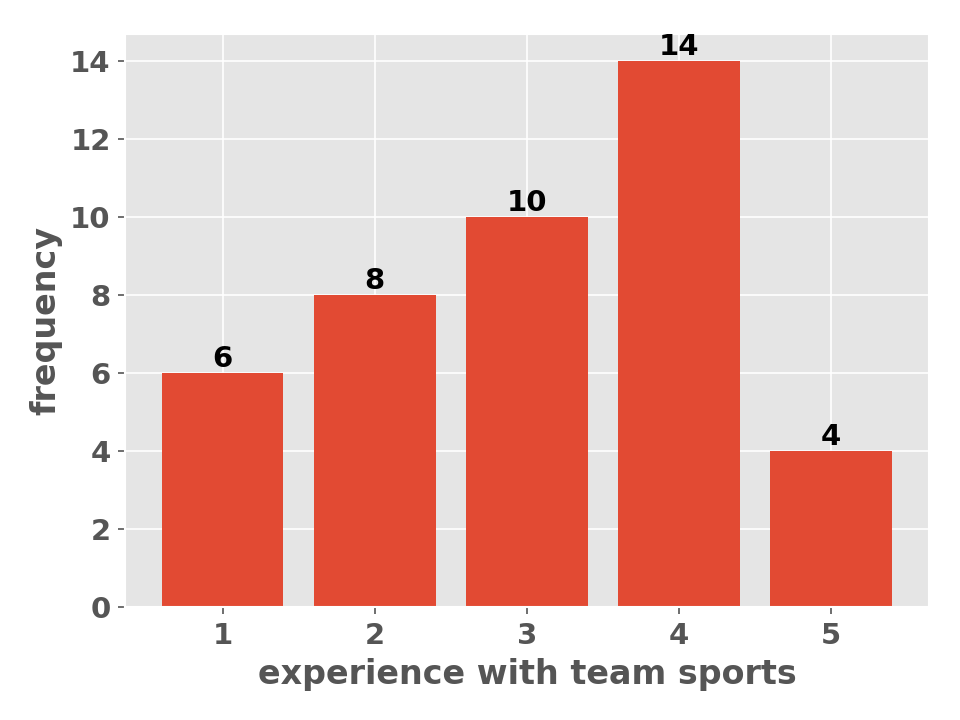}} \\\hline
\end{tabular}
\end{table*}
